\pdfoutput=1
\documentclass[11pt]{article}

\usepackage[final]{acl}
\usepackage{times}
\usepackage{latexsym}
\usepackage{amsmath}
\usepackage{graphicx}
\usepackage{listings}
\usepackage{hyperref}
\title{Different Corpora, Different Skills? A Paired Study of Small Language Models}

\author{
  \textbf{Edoardo Vaira} \and
  \textbf{Georgios Chavales} \and
  \textbf{Pierre Nugues}
\\
\\
Department of Computer Science\\
Faculty of Engineering\\
Lund University
\\
\href{mailto:edoardo.vaira@icloud.com}{edoardo.vaira@icloud.com} \\
\href{mailto:geochaval@gmail.com}{geochaval@gmail.com} \\
\href{mailto:pierre.nugues@cs.lth.se}{pierre.nugues@cs.lth.se}
}

\begin{document}
\maketitle

\begin{abstract}
Small language models are often trained under tight data budgets, where the choice of pretraining corpus may matter as much as model size or training recipe. We compare two contrasting corpora at matched 10-million- and 37-million-word budgets: the 1911 \textit{Encyclopaedia Britannica} and the BabyLM 2026 strict pretraining corpus. For each corpus, we train teacher models, distilled 58-million-parameter Baby-Llama students, and no-distillation student ablations, using the same model families, training recipes, and evaluation protocol across corpora. Across BLiMP, BLiMP Supplement, EWoK, Entity Tracking, and COMPS, the corpora produce different skill profiles rather than a single quality ranking. BabyLM 2026 is consistently stronger on linguistic acceptability benchmarks, while Britannica is often stronger on Entity Tracking, a synthetic state-tracking task. The contrast appears across teacher and student models and at both budgets, with the cleanest cross-seed replication in the distilled student comparison. Targeted controls show that the central contrast persists after matching training exposure and at a longer diagnostic context; direct teacher ensembling also does not reproduce the 10M Britannica student's Entity Tracking behavior. Entity Tracking magnitudes nevertheless remain sensitive to training dynamics. For this corpus pair, corpus choice can therefore produce different relative skill profiles at small scale, motivating multi-benchmark evaluation when comparing pretraining data.
\end{abstract}
 
\section{Introduction}

Frontier language models are trained on hundreds of billions to trillions of tokens \citep{chinchilla, llama, gpt3}. At that scale, corpus effects are entangled with large heterogeneous mixtures, filtering pipelines, and training recipes. In our setting, by contrast, a model's entire pretraining history comprises only tens of millions of words from one corpus. Recurring properties of that corpus---such as register, discourse structure, and sentence length---therefore make up a larger share of what the model observes and may have a more visible effect. Yet prior small-model work \citep{babylm, babyllama} often varies the corpus alongside the budget, tokenizer, or evaluation pipeline, making it difficult to attribute performance differences to the source data itself.

Our objective is to test whether changing the pretraining corpus under a matched procedure produces a uniform performance difference or changes the model's relative strengths across skills. We use small models as controlled experimental probes of this question, not as proposed deployment systems. We compare two contrasting corpora at matched 10-million- and 37-million-word budgets: the 1911 \textit{Encyclopaedia Britannica}, a single-source formal encyclopedia, and the BabyLM~2026 strict corpus, a heterogeneous mixture of child-directed speech, dialogue, simple Wikipedia, and transcribed speech. For each corpus, we first train larger \emph{teacher} language models directly on that corpus. We then train 58-million-parameter Baby-Llama \emph{student} models \citep{babyllama}, either with knowledge distillation (KD)---learning from both the text and the teachers' predictions---or without it. ``Matched'' means that corresponding models across the two corpora use the same architecture, tokenizer-training procedure, optimization recipe, and evaluation protocol; only the source data and the vocabulary learned from it differ. Consequently, the comparison matches source-word budget and procedure, but not the resulting vocabulary or token count.

We evaluate the models zero-shot, without task-specific training, on five benchmarks. The Benchmark of Linguistic Minimal Pairs (BLiMP) and its Supplement test grammatical acceptability \citep{blimp,babylm2026}; Elements of World Knowledge (EWoK) probes basic world knowledge \citep{ewok}; Entity Tracking tests state tracking and reference resolution in synthetic discourse \citep{entity_tracking}; and Conceptual Minimal Pair Sentences (COMPS) probes semantic property knowledge \citep{comps}. Across this suite, neither corpus dominates (Figure~\ref{fig:skill-profile}). BabyLM~2026 is consistently stronger on BLiMP and its Supplement, while Britannica is often stronger on Entity Tracking. This pattern appears in the teachers before distillation, persists in the students, and holds most clearly in the matched distilled-student comparisons at both budgets.

Our contributions are threefold. First, we provide a paired corpus comparison across two data budgets, two training seeds, teacher and student models, and distilled and non-distilled students, supplemented by a matched-exposure control. Second, we document a replicated cross-benchmark ranking reversal: BabyLM~2026 is stronger on linguistic acceptability, whereas Britannica is often stronger on synthetic state tracking. This shows, for this corpus pair, that changing the pretraining data can alter relative strengths rather than producing a uniform performance shift. Third, by observing the contrast in teachers before distillation and in students trained without it, while separately measuring corpus-dependent KD effects, we distinguish the central corpus result from a distillation artifact. Reporting only BLiMP would therefore produce a clean but partial ranking. This is especially relevant to cognitively motivated, data-limited modeling, where acquisition should be evaluated as a profile of emerging capabilities rather than a single aggregate score. Our code and configurations are available at \url{https://github.com/EdoardoVaira/babylm-corpora-skills}, and the Britannica corpus at \url{https://huggingface.co/datasets/EdoVaira/Encyclopedia-Britannica}.
 \section{Related work}

The effect of pretraining data composition on model behavior has been studied extensively at large scale. \citet{datacomp} introduce a controlled benchmark for dataset experiments with 412M--7B parameter models; \citet{fineweb} show that aggressive quality filtering yields large gains for billion-token models; \citet{pretrainersguide} systematically vary data age, domain coverage, quality, and toxicity; and \citet{datasurvey} survey the broader data-selection literature. These studies share a common finding---data quality and composition matter---but they operate at web-corpus scale with models far larger than those studied here, where corpus identity is entangled with filtering, mixing, and scaling choices.

At smaller scale, targeted data has supported strong results. \citet{tinystories} show that a constrained synthetic story corpus can train models below 10M parameters to generate coherent English. \citet{textbooks} train competitive models from 7B tokens of textbook-style data, and the Phi family \citep{phi3} extends this curated-data approach to billion-scale models. The BabyLM Challenge \citep{babylm} established matched-budget developmentally motivated pretraining as a testbed for small language models, and BabyLlama \citep{babyllama} showed that a 58M student can perform strongly after distillation from corpus-trained teachers. We use the BabyLM 2026 strict corpus \citep{babylm2026} as one side of our comparison. However, prior small-model work typically trains on a single corpus or varies the corpus alongside the budget, tokenizer, or evaluation pipeline, so corpus-specific effects are confounded with other design choices.

Knowledge distillation (KD) is a standard model-compression technique \citep{distill1, distill2}; in our setting, it is part of the BabyLlama-style training recipe rather than the main object of study. We adopt the BabyLlama distillation recipe and 58M student architecture \citep{babyllama}. Within each matched comparison, we substitute the source corpus while keeping the model family, tokenizer-training procedure, training recipe, and evaluation protocol fixed. Our contribution is not the generic observation that language models depend on their training corpus, which is long established. Rather, we use a paired small-model design to characterize a replicated reversal in the relative corpus ranking across acceptability and state-tracking evaluations.
 \section{Datasets}

We use two pretraining corpora at matched word budgets: the 1911 \textit{Encyclopaedia Britannica} and the BabyLM 2026 strict pretraining corpus \citep{babylm2026}. Table~\ref{tab:corpus-stats} summarizes their key properties.

\begin{table*}[tb]
\centering
\small
\begin{tabular}{lrrrr}
\hline
\textbf{Statistic} & \textbf{Brit. 10M} & \textbf{Brit. 37M} & \textbf{BabyLM 10M} & \textbf{BabyLM 37M} \\
\hline
Train documents & 9{,}181 & 34{,}295 & 1.17M & 4.30M \\
Mean document words & 1{,}089 & 1{,}079 & 8.6 & 8.6 \\
Median document words & 318 & 318 & 5 & 5 \\
Mean sentence words & 20.3 & 20.2 & 7.5 & 7.5 \\
Word types & 239k & 532k & 170k & 344k \\
Tokenizer tokens / word & 1.42 & 1.42 & 1.95 & 1.95 \\
\hline
\end{tabular}
\caption{Corpus properties for the matched 10M- and 37M-word training splits. Britannica is a single-source formal encyclopedia corpus; BabyLM~2026 is a heterogeneous mixture of child-directed speech, dialogue, simple Wikipedia, transcribed speech, and other text. Sentences are segmented on terminal punctuation. Token counts use each track's corpus-specific tokenizer and include document-boundary tokens.}
\label{tab:corpus-stats}
\end{table*}

\subsection{Corpora}

We chose Britannica because it is expert-edited with a consistent formal register, covers a wide range of topics, and provides a clear stylistic contrast to heterogeneous small-model pretraining corpora. We use the article volumes of the 11th edition (1911), scraped as volumes 1--28 and excluding the index volume. The source is digitally accessible, has a clear temporal boundary on its knowledge scope, and is public domain.

The BabyLM 2026 strict corpus provides the contrasting side of the comparison. It is a heterogeneous mixture of child-directed speech, dialogue, simple Wikipedia, transcribed speech, and other text. We use the larger \texttt{Strict} variant rather than \texttt{Strict-Small} because only \texttt{Strict} is large enough to support both a 10M-word and a 37M-word matched track from a single source.

The two corpora differ sharply in discourse structure. Britannica consists of long formal articles, for example:
\begin{quote}
\textit{``QUINSY, a common term for acute suppurative tonsillitis...''}
\end{quote}
BabyLM 2026 contains many short utterances and dialogue turns, for example:
\begin{quote}
\textit{``*MOT: does the monkey know how to drive the train?''}
\end{quote}
These differences in register, document length, sentence length, and topical continuity are the intended source of variation in our comparison.

\subsection{Preparation and splits}

We scrape Britannica articles from Wikisource, convert the scraped JSON to plain text, and apply a Britannica-specific cleanup step that normalizes Unicode, whitespace, quotes, dashes, and punctuation spacing, and removes bracketed numeric reference markers and \textit{q.v.}/\textit{cf.} parentheticals. For BabyLM~2026, we use the official Hugging Face release, whitespace-normalize each source row, and apply the same document-boundary token convention as for Britannica.

Both corpora are split into matched 10M-word and 37M-word training tracks, with approximately 2M-word and 1M-word validation splits respectively. Splits are formed by shuffling source documents with data-split seed 42, accumulating whole documents until the target word count is reached, and inserting explicit boundary tokens (\verb=<s>= and \verb=</s>=) between documents. Because whole documents are preserved, actual word counts can slightly exceed the nominal targets. The 10M training split is nested inside the 37M training split for each corpus; validation documents for the separate 10M-budget run can therefore appear in the 37M training split. Validation splits are used only for training monitoring; all reported scores come from external zero-shot benchmarks. Appendix~\ref{app:babylm-splits} reports the BabyLM 2026 split manifests, and Appendix~\ref{app:token-counts} reports token counts for all tracks.

\subsection{Tokenization and sequencing}

We train a separate 16{,}000-token byte-pair encoding (BPE) tokenizer on each track's training split, following the BabyLlama recipe \citep{babyllama}. The tokenizer-training procedure is fixed, but vocabularies and token counts are corpus-specific; we therefore treat this as a matched-procedure comparison rather than a shared-vocabulary study. Because training uses fixed 128-token slices, matched source-word budgets do not imply matched token budgets. In these splits, BabyLM 2026 yields about 1.37$\times$ as many BPE training tokens as Britannica at both budgets, reflecting both corpus-specific BPE segmentation and BabyLM's many short source segments, which introduce far more document-boundary markers.

The tokenized text is segmented into 128-token slices. Longer documents span multiple slices, while shorter documents can share a slice with neighbors. During training, the slice start is randomly offset within a 128-token window; during validation, slices are deterministic and contiguous. Any final incomplete slice is discarded rather than padded.
 \section{Model architecture and training}
\label{sec:architecture}

To separate corpus effects from artifacts of a single architecture or distillation pipeline, we train three kinds of models per corpus: (i) larger teacher models trained directly on the corpus, (ii) 58M-parameter Baby-Llama students distilled from an ensemble of corpus-matched teachers, and (iii) 58M students trained without distillation on the same data. All models are trained from scratch on 128-token sequences with the corpus-specific tokenizer for that track. Although the architectures support a maximum context of 256 tokens, we use this extended capacity only for the diagnostic context-length check in Section~\ref{sec:eval}. We run the full matched sweep with training seeds 2023 and 43, separate from the data-split seed, and evaluate all models zero-shot. Absolute-score tables use seed 2023 for readability, while Table~\ref{tab:corpusdelta} reports the paired corpus effects for both seeds and their across-seed variability.

\subsection{Teacher models}

We train multiple teacher architectures on each corpus to verify that corpus effects are not specific to a single architecture. At the 10M budget, we use three teachers (Llama-360M, GPT-2-705M, GPT-J-705M); at 37M, we use two (Llama-360M, GPT-2-705M). Table~\ref{tab:teacher-hp} summarizes the architectures and training recipes. All teachers use a cosine learning-rate schedule with 300 warm-up steps; validation loss is computed after each epoch on a fixed random subset of 8{,}192 sequences.

\begin{table}[tb]
\centering
\small
\resizebox{\columnwidth}{!}{
\begin{tabular}{lrrr}
\hline
 & \textbf{Llama} & \textbf{GPT-2} & \textbf{GPT-J} \\
\hline
Layers & 24 & 24 & 24 \\
Attention heads & 8 & 16 & 16 \\
Hidden dimension & 1024 & 1536 & 1536 \\
Intermediate dimension & 3072 & 6144 & 6144 \\
Parameters & 360M & 705M & 705M \\
Epochs & 4 & 6 & 4 \\
Effective batch size & 128 & 256 & 128 \\
Peak learning rate & $3\!\times\!10^{-4}$ & $2.5\!\times\!10^{-4}$ & $2.5\!\times\!10^{-4}$ \\
\hline
\end{tabular}}
\caption{Teacher architectures and training recipes. All teachers share a cosine learning-rate schedule with 300 warm-up steps. GPT-J is trained only at the 10M budget.}
\label{tab:teacher-hp}
\end{table}

\subsection{Student models}

All 58M students share a single Llama-based architecture: 16 transformer layers, 8 attention heads, hidden dimension 512, and intermediate dimension 1024 ($\sim$58M parameters). Each is trained from scratch for 6 epochs with effective batch size 32, peak learning rate $2.5 \times 10^{-4}$, a cosine schedule with 200 warm-up steps, and weight decay 0.1. As with the teacher runs, validation loss is computed after each epoch on a fixed subset of 8{,}192 validation sequences.

\paragraph{Distilled students.}
We follow the BabyLlama recipe \citep{babyllama, distill1, distill2}. The student is trained against both next-token labels and soft targets obtained by averaging the raw logits of two corpus-matched teachers (Llama + GPT-2 by default, or Llama + GPT-J for the alternate-teacher 10M variant), then applying temperature $T = 2$. Teacher logits are computed on the same training batches seen by the student. The loss is
\[
L = \alpha L_{\text{CE}} + (1 - \alpha) T^2 L_{\text{KL}},
\]
with $\alpha = 0.5$, where $L_{\text{CE}}$ is cross-entropy on the next-token labels and $L_{\text{KL}}$ is the KL divergence between the student's softened distribution and the softened distribution of the averaged teacher logits. In the implementation, the model-provided cross-entropy is averaged over non-masked tokens, while the KL term uses PyTorch's \texttt{batchmean} reduction (summed over token positions and divided by batch size).

\paragraph{No-distillation baseline.}
At both word budgets, we additionally train a student with the same architecture, optimizer, schedule, and number of epochs, but using plain cross-entropy on next-token labels with no teacher signal. This lets us separate corpus effects from distillation effects.

\subsection{Compute}

All runs use a single NVIDIA A40 graphics processing unit (GPU) with 16-bit floating-point (fp16) training. At 10M words, training takes roughly 1--2 hours for a Llama teacher, 3--4.5 hours for a GPT-style teacher, and 4--6 hours for a distilled 58M student; no-distillation students omit teacher forward passes and take roughly 0.5--0.8 hours. At 37M words, training scales to roughly 4--6 hours for a Llama teacher, 11--16 hours for a GPT-2 teacher, about 15--20 hours for a distilled 58M student, and 2--2.5 hours for a no-distillation student.
 \section{Experimental results}

\subsection{Evaluation suite}
\label{sec:eval}

We evaluate every trained model on a zero-shot suite drawn from the BabyLM 2026 strict evaluation pipeline \citep{babylm2026}, which extends the BLiMP-only protocol used in earlier BabyLM Challenge editions. The main text focuses on five metrics:
\begin{itemize}
    \item \textbf{BLiMP} \citep{blimp} -- 67 minimal-pair tasks targeting syntax, semantics, and morphosyntax.
    \item \textbf{BLiMP Supplement} \citep{babylm2026} -- an extension of BLiMP with additional and more diverse minimal-pair tasks.
    \item \textbf{EWoK} \citep{ewok} -- minimal-pair world-knowledge probes.
    \item \textbf{Entity Tracking} \citep{entity_tracking} -- a synthetic state-tracking benchmark in which a context describes a sequence of operations on entities (e.g.\ boxes containing objects), and the model must select the continuation that correctly describes the resulting state.
    \item \textbf{COMPS} \citep{comps} -- minimal-pair semantic plausibility.
\end{itemize}

For all benchmarks, higher scores indicate better performance. Because our main protocol uses 128-token training sequences and a tokenizer-declared context window of 128 tokens, we apply left-truncation to the zero-shot evaluation inputs. To check whether the Entity Tracking results are driven by left-truncation (since prompts at higher operation counts can exceed 128 tokens), we additionally re-evaluate all models at 256-token context---the maximum supported by the models' position embeddings.\footnote{The 256-token setting is diagnostic rather than the main protocol: all models were trained on 128-token slices. The Llama-based models use rotary position embeddings (RoPE), while GPT-2 teachers use learned absolute position embeddings; some GPT-2 teacher rows, especially at 10M, are therefore sensitive to positions 128--255.} The full 256-token Entity Tracking results are reported in Appendix~\ref{app:context-robustness}; the Britannica-vs-BabyLM ranking and the student-vs-ensemble patterns reported below hold at both context lengths. We also run the BabyLM reading benchmark, which reports correlations against eye-tracking (Eye) and self-paced reading (SPR) data; scores are low overall and do not change the main corpus-tradeoff conclusion, so we report them only in Appendix~\ref{app:full-results}.

\subsection{Main results: different corpora, different skills}
\label{sec:main-results}

\begin{figure*}[t]
\centering
\includegraphics[width=\textwidth]{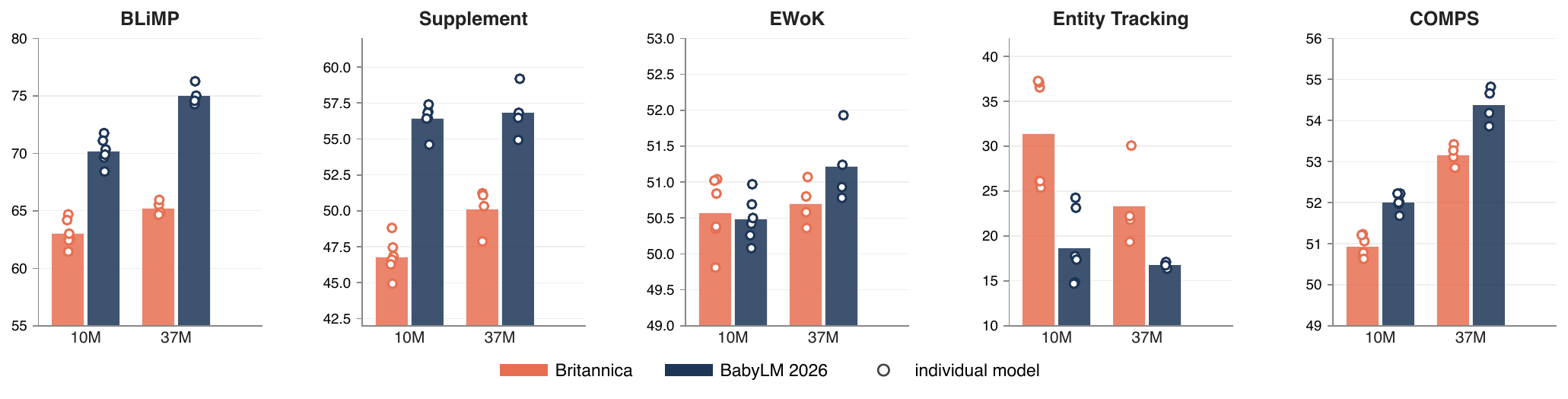}
\caption{Skill profiles of the seed-2023 experimental sweep on five zero-shot metrics. Each panel shows one metric; bars give the mean accuracy per (corpus, budget) combination, and circles mark individual models (3 teachers + 3 students at 10M; 2 teachers + 2 students at 37M). The Britannica corpus produces higher Entity Tracking in the main matched comparisons; the BabyLM~2026 corpus produces higher BLiMP, BLiMP Supplement, and (slightly) COMPS; EWoK is essentially tied. Table~\ref{tab:corpusdelta} reports the paired effects for both training seeds.}
\label{fig:skill-profile}
\end{figure*}

Figure~\ref{fig:skill-profile} gives the seed-2023 skill profiles. We next test whether the paired corpus effects replicate under seed 43. Table~\ref{tab:corpusdelta} reports BabyLM~2026 minus Britannica for every matched model family under both seeds, so positive BLiMP/Supplement values indicate a BabyLM advantage, while negative Entity Tracking values indicate a Britannica advantage.

\begin{table*}[tb]
\centering
\resizebox{\textwidth}{!}{
\begin{tabular}{llrrrrrr}
\hline
& & \multicolumn{2}{c}{\textbf{$\Delta$BLiMP}} & \multicolumn{2}{c}{\textbf{$\Delta$Supp}} & \multicolumn{2}{c}{\textbf{$\Delta$Entity}} \\
\textbf{Budget} & \textbf{Matched model family} & \textbf{s2023} & \textbf{s43} & \textbf{s2023} & \textbf{s43} & \textbf{s2023} & \textbf{s43} \\
\hline
10M & Llama-360M teacher & $+$8.19 & $+$7.97 & $+$7.65 & $+$10.46 & $-$8.38 & $-$8.05 \\
10M & GPT-2-705M teacher & $+$7.73 & $+$8.19 & $+$7.78 & $+$5.21 & $-$8.02 & $-$16.49 \\
10M & GPT-J-705M teacher & $+$5.99 & $+$7.22 & $+$11.96 & $+$5.53 & $-$12.29 & $-$4.59 \\
10M & Baby-Llama-58M student (KD, Llama+GPT-2) & $+$7.07 & $+$8.19 & $+$10.23 & $+$11.44 & $-$22.27 & $-$17.00 \\
10M & Baby-Llama-58M student (KD, Llama+GPT-J) & $+$6.91 & $+$6.41 & $+$10.15 & $+$11.75 & $-$22.58 & $-$5.49 \\
10M & Baby-Llama-58M student (no KD) & $+$6.88 & $+$8.97 & $+$9.95 & $+$11.09 & $-$2.97 & $+$6.49 \\
37M & Llama-360M teacher & $+$8.76 & $+$8.83 & $+$3.71 & $+$4.72 & $-$2.34 & $-$1.52 \\
37M & GPT-2-705M teacher & $+$10.16 & $+$9.61 & $+$8.87 & $+$4.90 & $-$13.70 & $+$7.44 \\
37M & Baby-Llama-58M student (KD, Llama+GPT-2) & $+$10.32 & $+$10.79 & $+$5.74 & $+$8.42 & $-$4.70 & $-$5.06 \\
37M & Baby-Llama-58M student (no KD) & $+$9.91 & $+$9.66 & $+$8.60 & $+$5.36 & $-$5.49 & $-$12.23 \\
\hline
\end{tabular}}
\caption{Paired corpus effects under both training seeds. Deltas are BabyLM~2026 minus Britannica for the same model family and source-word budget. Positive BLiMP/Supplement values favor BabyLM~2026; negative Entity Tracking values favor Britannica.}
\label{tab:corpusdelta}
\end{table*}

All 20 matched BLiMP comparisons favor BabyLM~2026 across the two seeds. Eighteen of 20 Entity Tracking comparisons favor Britannica; the two reversals are the seed-43 37M GPT-2 teacher and 10M no-KD student. Every matched KD-student comparison preserves both directions. For the default Llama+GPT-2 KD students, the paired effects computed from the unrounded benchmark-subtask accuracies (mean $\pm$ sample standard deviation over two seeds) are $+7.64\pm0.79$ BLiMP, $+10.83\pm0.86$ Supplement, and $-19.63\pm3.73$ Entity Tracking at 10M; at 37M they are $+10.55\pm0.33$, $+7.08\pm1.89$, and $-4.88\pm0.25$, respectively. Because these summaries use unrounded accuracies, they can differ by 0.01 from averages reconstructed from the rounded table entries. These standard deviations summarize two runs and therefore provide only a limited estimate of training variability. Separately, paired hierarchical-bootstrap 95\% confidence intervals over benchmark-subtask groups and paired items exclude zero for all three metrics at both budgets and both seeds (Appendix~\ref{app:bootstrap-uncertainty}); these intervals quantify evaluation-set uncertainty conditional on each checkpoint, not training variability. Appendix~\ref{app:full-results} reports the seed-2023 absolute scores.

Because the BabyLM split yields 1.37$\times$ as many BPE tokens per matched word budget, we also run a 10M matched-exposure control covering the Llama and GPT-2 teachers used by the default KD student, that KD student, and the no-KD student. We retain the BabyLM split and tokenizer but cap each model at its Britannica counterpart's optimizer-step count, thereby matching schedule length and approximately matching 128-token sequence presentations. BabyLM remains ahead on BLiMP and Supplement in all four controlled models. Britannica remains ahead on Entity Tracking in both controlled teachers and the KD student; the no-KD comparison reverses to a small 1.28-point BabyLM advantage, consistent with that row's seed sensitivity above. Appendix~\ref{app:matched-exposure} gives the design and results. This control matches total training exposure, not the number of unique tokens in the two corpora.

Tables~\ref{tab:main10m} and \ref{tab:main37m} zoom in on the 58M students at both budgets.

\begin{table*}[tb]
\centering
\begin{tabular}{lrrrrr}
\hline
\textbf{Model} & \textbf{BLiMP} & \textbf{Supp} & \textbf{EWoK} & \textbf{Entity} & \textbf{COMPS} \\
\hline
\multicolumn{6}{l}{\emph{Britannica 10M}} \\
\quad Baby-Llama-58M (KD, Llama+GPT-2) & 64.69 & 46.60 & 49.81 & 37.10 & 51.23 \\
\quad Baby-Llama-58M (KD, Llama+GPT-J) & 64.19 & 46.27 & \textbf{51.02} & \textbf{37.26} & 51.21 \\
\quad Baby-Llama-58M (no KD) & 63.01 & 47.45 & 50.84 & 26.11 & 50.63 \\
\multicolumn{6}{l}{\emph{BabyLM 2026 10M}} \\
\quad Baby-Llama-58M (KD, Llama+GPT-2) & \textbf{71.76} & 56.83 & 50.08 & 14.83 & 52.00 \\
\quad Baby-Llama-58M (KD, Llama+GPT-J) & 71.10 & 56.42 & 50.26 & 14.68 & \textbf{52.22} \\
\quad Baby-Llama-58M (no KD) & 69.89 & \textbf{57.40} & 50.97 & 23.14 & 51.68 \\
\hline
\end{tabular}
\caption{Seed-2023 zero-shot evaluation of 58M students at matched 10M-word budget on the Britannica and BabyLM~2026 corpora. KD labels name the two-teacher ensemble used for distillation; ``no KD'' is the matched non-distilled baseline. Bold marks the best entry per metric across both corpora.}
\label{tab:main10m}
\end{table*}

\begin{table*}[tb]
\centering
\begin{tabular}{llrrrrr}
\hline
\textbf{Corpus} & \textbf{Model} & \textbf{BLiMP} & \textbf{Supp} & \textbf{EWoK} & \textbf{Entity} & \textbf{COMPS} \\
\hline
Britannica 37M & KD, Llama+GPT-2 & 65.95 & 51.08 & 50.36 & 21.80 & 53.42 \\
Britannica 37M & no KD & 64.66 & 47.87 & 50.58 & \textbf{22.22} & 53.27 \\
BabyLM 2026 37M & KD, Llama+GPT-2 & \textbf{76.27} & \textbf{56.82} & \textbf{51.24} & 17.10 & \textbf{54.66} \\
BabyLM 2026 37M & no KD & 74.57 & 56.47 & 50.93 & 16.73 & 54.18 \\
\hline
\end{tabular}
\caption{Seed-2023 zero-shot evaluation of 58M Baby-Llama students at the larger 37M-word budget. KD rows use the Llama+GPT-2 two-teacher distillation ensemble; no-KD rows use the same architecture and training schedule without teacher logits. Bold marks the best entry per metric across both corpora.}
\label{tab:main37m}
\end{table*}

Three further points are visible in these tables. First, the 58M students are competitive with the larger teachers on BLiMP: at 10M, the no-KD student exceeds all three Britannica teachers (63.01 vs.\ 61.45--62.61) and is within 0.45 points of the best BabyLM~2026 teacher (69.89 vs.\ 70.34). Thus, the corpus comparison remains informative even though the students are substantially smaller than the teachers.

Second, within the two 10M KD configurations, swapping GPT-J for GPT-2 as the second teacher changes student BLiMP by less than 0.7 points on either corpus and does not alter the BLiMP/Supplement versus Entity Tracking pattern. In this comparison, the corpus-associated differences are larger than those between the two teacher pairs.

Third, the cross-over persists at 37M. In the KD students, BabyLM~2026 gains a further $\sim$5 BLiMP points over its 10M student (71.76 $\rightarrow$ 76.27) while Britannica gains $\sim$1 point (64.69 $\rightarrow$ 65.95). On Entity Tracking, the Britannica KD student drops from 37.10 to 21.80, while the BabyLM~2026 KD student improves slightly (14.83 $\rightarrow$ 17.10), leaving Britannica ahead by $\sim$5 points. The 37M no-KD students show the same corpus direction (22.22 vs.\ 16.73 on Entity Tracking), confirming that the larger-budget contrast is not created by distillation alone. EWoK is essentially tied across corpora and budgets; COMPS shows a small consistent BabyLM~2026 advantage.

\subsection{Ablations and analysis}
\label{sec:ablations}

\paragraph{Effect of distillation.}
Table~\ref{tab:nokd} reports the paired effect of distillation: KD score minus no-KD score on the same corpus and budget.

\begin{table*}[tb]
\centering
\begin{tabular}{lrrrrr}
\hline
\textbf{Corpus} & $\Delta$\textbf{BLiMP} & $\Delta$\textbf{Supp} & $\Delta$\textbf{EWoK} & $\Delta$\textbf{Entity} & $\Delta$\textbf{COMPS} \\
\hline
Britannica 10M & +1.68 & -0.85 & -1.03 & +10.99 & +0.60 \\
BabyLM 2026 10M & +1.87 & -0.57 & -0.89 & -8.31 & +0.32 \\
Britannica 37M & +1.29 & +3.21 & -0.22 & -0.42 & +0.15 \\
BabyLM 2026 37M & +1.70 & +0.35 & +0.31 & +0.37 & +0.48 \\
\hline
\end{tabular}
\caption{Effect of knowledge distillation on the 58M student, expressed as KD score minus no-KD score on the same corpus and word budget. Positive values indicate that distillation helps.}
\label{tab:nokd}
\end{table*}

For the default Llama+GPT-2 KD rows at 10M, distillation improves BLiMP by $\sim$1.7--1.9 points on both corpora and helps COMPS slightly. At 10M it is also mildly negative on Supplement and EWoK regardless of corpus. The striking result is Entity Tracking, where the effect is large but opposite-signed: $+$10.99 on Britannica versus $-$8.31 on BabyLM~2026. Distillation therefore amplifies the corpus-level Entity Tracking contrast rather than reducing it: a $\sim$3-point Britannica advantage between the no-KD students (26.11 vs.\ 23.14) grows to a 22-point gap once distillation is applied (37.10 vs.\ 14.83). At 37M, the KD effect on Entity Tracking is much smaller on both corpora ($-$0.42 on Britannica, $+$0.37 on BabyLM~2026), so the dramatic Entity Tracking amplification is specific to the 10M setting.

On BabyLM~2026, the no-KD baseline scores 23.14 on Entity Tracking (above the 20\% chance level), but KD pulls the student down to 14.83 (below chance), suggesting that the distillation signal can move the student's preferences away from the correct continuation on this corpus. On Britannica, by contrast, the KD student reaches 37.10, roughly 11 points higher than either individual teacher (Llama 26.13, GPT-2 25.38).

\paragraph{Ensemble baseline and the role of data budget.}
A natural hypothesis is that the Britannica KD student simply inherits the predictions of the teacher ensemble: even when each teacher individually scores around 26, their logit mean might concentrate more reliably on the correct option. We test this directly by scoring Entity Tracking with the per-token logit mean of each KD student's two corpus-matched teachers, using the same evaluation pipeline. Table~\ref{tab:ensemble} reports the result.

\begin{table*}[tb]
\centering
\resizebox{\textwidth}{!}{
\begin{tabular}{rllrrr}
\hline
\textbf{Seed} & \textbf{Budget} & \textbf{Teacher pair / context} & \textbf{Teacher ensemble} & \textbf{KD student} & \textbf{Student $-$ Ensemble} \\
\hline
2023 & 10M & Llama+GPT-2 @ 128 tokens & 25.15 & 37.10 & $+$11.95 \\
43 & 10M & Llama+GPT-2 @ 128 tokens & 29.15 & 35.58 & $+$6.43 \\
2023 & 10M & Llama+GPT-2 @ 256 tokens & 15.83 & 38.96 & $+$23.13 \\
2023 & 10M & Llama+GPT-J @ 128 tokens & 31.29 & 37.26 & $+$5.97 \\
2023 & 10M & Llama+GPT-J @ 256 tokens & 26.94 & 37.71 & $+$10.77 \\
\hline
2023 & 37M & Llama+GPT-2 @ 128 tokens & 23.85 & 21.80 & $-$2.05 \\
2023 & 37M & Llama+GPT-2 @ 256 tokens & 27.02 & 22.05 & $-$4.97 \\
\hline
\end{tabular}
}
\caption{Direct ensemble comparison for the Britannica Entity Tracking models. All entries are macro averages: accuracy is first computed for each benchmark subtask (UID group) and then averaged. For the ensemble, each candidate completion is scored by the per-token logit mean of the two Britannica-trained teachers used during KD training. Main rows use seed 2023; the seed-43 row repeats the default 10M Llama+GPT-2 check. At 10M, the student exceeds the ensemble in every configuration; at 37M, the ensemble exceeds the student at both context lengths.}
\label{tab:ensemble}
\end{table*}

At 10M, the ensemble scores well below the KD student in every configuration: the Llama+GPT-2 ensemble reaches 25.15 versus 37.10 for the student (a $+$11.95 gap), and the Llama+GPT-J ensemble reaches 31.29 versus 37.26 ($+$5.97). Repeating the default Llama+GPT-2 check under seed~43 gives the same direction (29.15 versus 35.58, a $+$6.43 gap). Direct logit averaging at evaluation time therefore does not explain the behavior learned by the KD student; the simple ``KD inherits the ensemble'' account is insufficient.

The picture reverses at 37M. The Llama+GPT-2 ensemble scores 23.85 at 128-token context and 27.02 at 256-token context, exceeding the KD student (21.80 and 22.05). The student-over-ensemble lift observed at 10M is therefore budget-specific rather than a generic property of Britannica~+~KD.

The 37M no-KD rows confirm a $\sim$5.5-point Britannica Entity Tracking advantage without distillation at the larger budget. Unlike at 10M, where KD amplifies the Entity Tracking gap from $\sim$3 to $\sim$22 points, at 37M the gap is comparable with and without KD (22.22 vs.\ 16.73 without KD; 21.80 vs.\ 17.10 with KD). The KD advantage on BLiMP also narrows slightly at 37M: the no-KD students score 74.57 (BabyLM) and 64.66 (Britannica), versus KD students at 76.27 and 65.95, yielding KD lifts of $+$1.70 and $+$1.29. These are comparable to the 10M lifts ($+$1.87 and $+$1.68), so the KD benefit on linguistic acceptability is stable across budgets, while the dramatic Entity Tracking amplification is specific to 10M.

More broadly, comparing the 10M and 37M students across both KD and no-KD conditions, the larger-budget setup improves BLiMP and COMPS on both corpora. The main skill-profile contrast is preserved at the larger budget and is visible in the no-KD baseline, supporting an association with the pretraining data rather than an effect created solely by distillation.

\paragraph{What Entity Tracking measures.}
Because Entity Tracking is the strongest single metric favoring Britannica, it is worth being precise about what it tests. The benchmark measures state-tracking and reference resolution over synthetic discourse---not encyclopedic factual knowledge. The Britannica advantage should not be read as stronger world knowledge. A more plausible interpretation is that the structural regularities of formal encyclopedia prose (long descriptive passages, frequent third-person reference, stable topical focus) better match the discourse-level patterns probed by the benchmark. We do not have a mechanistic account and treat this as an observation worth further investigation.
 \section{Conclusion}

Our paired comparison reveals a replicated ranking inversion rather than a single corpus ordering: BabyLM~2026 is stronger on linguistic acceptability, while Britannica is often stronger on Entity Tracking. The contrast appears in teachers and distilled students, at both budgets, and---for both teachers and the distilled student---after matching total training exposure at 10M.

Distillation improves BLiMP but is not uniformly beneficial, and the no-distillation baselines show that it does not create the broad corpus contrast. Reporting only BLiMP would therefore favor BabyLM~2026 throughout, whereas Entity Tracking would often favor Britannica. Multi-benchmark evaluation is necessary to reveal these different skill profiles. For cognitively motivated small-model research, limited pretraining experience should therefore be evaluated through the profile of capabilities that emerges, not only a single aggregate score.

Entity Tracking measures synthetic state tracking, not encyclopedic knowledge. With only two corpora, we cannot identify whether register, discourse structure, repetition, or another property causes the differences; targeted corpus manipulations are needed. All code and configurations are available at \url{https://github.com/EdoardoVaira/babylm-corpora-skills}; the Britannica corpus is at \url{https://huggingface.co/datasets/EdoVaira/Encyclopedia-Britannica}.

\section*{Limitations}

Our comparison is between two corpora; with $n=2$, the relationship between corpus structure and skill profile is suggestive rather than conclusive. The student-side ablations use a single 58M-parameter architecture, a single distillation recipe inherited from BabyLlama, and zero-shot evaluation; whether the skill-profile contrast carries over to other student sizes, distillation setups, or evaluation paradigms is open. The full matched sweep uses two training seeds. Table~\ref{tab:corpusdelta} reports both runs and their paired effects; with only two seeds, the reported sample standard deviations remain limited estimates of training variability. The matched distilled-student comparisons preserve the acceptability-versus-Entity-Tracking direction in every case, while two seed-43 Entity Tracking rows outside the KD comparisons reverse direction, so we treat Entity Tracking magnitudes as seed-sensitive. The no-KD baseline was run at both 10M and 37M; at 37M the Entity Tracking corpus gap is similar with and without KD ($\sim$5.5-point Britannica advantage), suggesting that the dramatic 10M KD amplification (Section~\ref{sec:ablations}) is budget-specific rather than a generic property of distillation. Source-word matching does not imply token-budget matching under corpus-specific BPE tokenization; Appendix~\ref{app:token-counts} reports the difference. Our seed-2023 matched-exposure control equalizes optimizer steps and total sequence presentations at 10M but retains the larger unique BabyLM corpus; it is therefore not a trim-matched corpus study, and its training variability is not separately estimated. The Entity Tracking benchmark is synthetic, and we caution against reading its scores as evidence about open-world knowledge.

\section*{Ethics Statement}

The 1911 \textit{Encyclopaedia Britannica} contains historically dated viewpoints and outdated factual content; models trained on it may reproduce these characteristics, and the comparison reported here should not be taken as endorsement of the corpus as training data for downstream user-facing applications. The small models studied here are not intended for deployment as general-purpose language models. The skill-profile finding is observational; we caution against using it to motivate corpus selection in safety-relevant applications without further targeted evaluation.

\section*{Acknowledgments}

This work was partially supported by \textit{Vetenskapsrådet}, the Swedish Research Council, registration number 2021-04533.
 
\bibliography{custom}

\appendix
\section{Full Results and Split Manifests}
\label{app:full-results}

The main text reports paired corpus deltas and the 58M student results. Tables~\ref{tab:app10m} and \ref{tab:app37m} give the corresponding absolute zero-shot scores for the main teacher, KD-student, and no-KD student models. Reading columns report correlation scores for eye-tracking (Eye) and self-paced reading (SPR); reading scores were not collected for the 37M no-KD rows.

\begin{table*}[tb]
\centering
\resizebox{\textwidth}{!}{
\begin{tabular}{llrrrrrrr}
\hline
\textbf{Corpus} & \textbf{Model} & \textbf{BLiMP} & \textbf{Supp} & \textbf{EWoK} & \textbf{Entity} & \textbf{COMPS} & \textbf{Eye} & \textbf{SPR} \\
\hline
Britannica 10M & Llama-360M teacher & 61.45 & 48.81 & 50.35 & 26.13 & 50.69 & 0.35 & 0.01 \\
Britannica 10M & GPT-2-705M teacher & 62.61 & 46.83 & 51.04 & 25.38 & 51.06 & 0.12 & 0.01 \\
Britannica 10M & GPT-J-705M teacher & 62.43 & 44.93 & 50.38 & 36.54 & 50.78 & 0.25 & 0.03 \\
Britannica 10M & Baby-Llama-58M student (KD, Llama+GPT-2) & 64.69 & 46.60 & 49.81 & 37.10 & 51.23 & 0.13 & 0.01 \\
Britannica 10M & Baby-Llama-58M student (KD, Llama+GPT-J) & 64.19 & 46.27 & 51.02 & 37.26 & 51.21 & 0.17 & 0.00 \\
Britannica 10M & Baby-Llama-58M student (no KD) & 63.01 & 47.45 & 50.84 & 26.11 & 50.63 & 0.26 & 0.07 \\
BabyLM 2026 10M & Llama-360M teacher & 69.64 & 56.46 & 50.42 & 17.75 & 51.97 & 0.17 & 0.00 \\
BabyLM 2026 10M & GPT-2-705M teacher & 70.34 & 54.61 & 50.50 & 17.36 & 52.22 & 0.18 & 0.03 \\
BabyLM 2026 10M & GPT-J-705M teacher & 68.42 & 56.89 & 50.69 & 24.25 & 52.02 & 0.34 & 0.02 \\
BabyLM 2026 10M & Baby-Llama-58M student (KD, Llama+GPT-2) & 71.76 & 56.83 & 50.08 & 14.83 & 52.00 & 0.14 & 0.02 \\
BabyLM 2026 10M & Baby-Llama-58M student (KD, Llama+GPT-J) & 71.10 & 56.42 & 50.26 & 14.68 & 52.22 & 0.15 & 0.02 \\
BabyLM 2026 10M & Baby-Llama-58M student (no KD) & 69.89 & 57.40 & 50.97 & 23.14 & 51.68 & 0.10 & 0.08 \\
\hline
\end{tabular}}
\caption{Full zero-shot results for the matched 10M-word corpus comparisons. KD student labels name the two-teacher ensemble used for distillation.}
\label{tab:app10m}
\end{table*}

\begin{table*}[tb]
\centering
\resizebox{\textwidth}{!}{
\begin{tabular}{llrrrrrrr}
\hline
\textbf{Corpus} & \textbf{Model} & \textbf{BLiMP} & \textbf{Supp} & \textbf{EWoK} & \textbf{Entity} & \textbf{COMPS} & \textbf{Eye} & \textbf{SPR} \\
\hline
Britannica 37M & Llama-360M teacher & 65.52 & 51.22 & 50.80 & 19.34 & 53.11 & 0.41 & 0.02 \\
Britannica 37M & GPT-2-705M teacher & 64.86 & 50.32 & 51.07 & 30.06 & 52.85 & 0.39 & 0.01 \\
Britannica 37M & Baby-Llama-58M student (KD, Llama+GPT-2) & 65.95 & 51.08 & 50.36 & 21.80 & 53.42 & 0.31 & 0.02 \\
Britannica 37M & Baby-Llama-58M student (no KD) & 64.66 & 47.87 & 50.58 & 22.22 & 53.27 & -- & -- \\
BabyLM 2026 37M & Llama-360M teacher & 74.28 & 54.93 & 50.78 & 17.00 & 53.86 & 0.24 & 0.01 \\
BabyLM 2026 37M & GPT-2-705M teacher & 75.02 & 59.19 & 51.93 & 16.36 & 54.82 & 0.19 & 0.02 \\
BabyLM 2026 37M & Baby-Llama-58M student (KD, Llama+GPT-2) & 76.27 & 56.82 & 51.24 & 17.10 & 54.66 & 0.18 & 0.01 \\
BabyLM 2026 37M & Baby-Llama-58M student (no KD) & 74.57 & 56.47 & 50.93 & 16.73 & 54.18 & -- & -- \\
\hline
\end{tabular}}
\caption{Full zero-shot results for the matched 37M-word corpus comparisons. Reading scores for the 37M no-KD runs were not collected; these models were evaluated only on the five core metrics.}
\label{tab:app37m}
\end{table*}

\section{Entity Tracking at 256-Token Context}
\label{app:context-robustness}

Entity Tracking prompts vary in length depending on the number of operations: at 4--5 ops, prompts can reach $\sim$200--250 tokens and so are left-truncated when scored against the models' 128-token tokenizer setting. To check whether the corpus and student-vs-teacher patterns reported in Sections~\ref{sec:main-results} and~\ref{sec:ablations} are artifacts of this truncation, we re-evaluated every Entity Tracking row at 256-token context---the maximum supported by the models' position embeddings. Table~\ref{tab:app-et256} reports the comparison.

\begin{table*}[tb]
\centering
\begin{tabular}{llrrr}
\hline
\textbf{Corpus} & \textbf{Model} & \textbf{ET @ 128} & \textbf{ET @ 256} & \textbf{$\Delta$} \\
\hline
Britannica 10M & Llama-360M teacher & 26.13 & 23.00 & $-$3.13 \\
Britannica 10M & GPT-2-705M teacher & 25.38 & 13.48 & $-$11.90 \\
Britannica 10M & GPT-J-705M teacher & 36.54 & 32.08 & $-$4.46 \\
Britannica 10M & Baby-Llama-58M (Llama+GPT-2) & 37.10 & 38.96 & $+$1.86 \\
Britannica 10M & Baby-Llama-58M (Llama+GPT-J) & 37.26 & 37.71 & $+$0.45 \\
Britannica 10M & Baby-Llama-58M (no KD) & 26.11 & 23.34 & $-$2.77 \\
BabyLM 2026 10M & Llama-360M teacher & 17.75 & 16.54 & $-$1.21 \\
BabyLM 2026 10M & GPT-2-705M teacher & 17.36 & 13.21 & $-$4.15 \\
BabyLM 2026 10M & GPT-J-705M teacher & 24.25 & 25.30 & $+$1.05 \\
BabyLM 2026 10M & Baby-Llama-58M (Llama+GPT-2) & 14.83 & 15.48 & $+$0.65 \\
BabyLM 2026 10M & Baby-Llama-58M (Llama+GPT-J) & 14.68 & 14.74 & $+$0.06 \\
BabyLM 2026 10M & Baby-Llama-58M (no KD) & 23.14 & 18.92 & $-$4.22 \\
Britannica 37M & Llama-360M teacher & 19.34 & 22.91 & $+$3.57 \\
Britannica 37M & GPT-2-705M teacher & 30.06 & 30.30 & $+$0.24 \\
Britannica 37M & Baby-Llama-58M (Llama+GPT-2) & 21.80 & 22.05 & $+$0.25 \\
Britannica 37M & Baby-Llama-58M (no KD) & 22.22 & 22.59 & $+$0.37 \\
BabyLM 2026 37M & Llama-360M teacher & 17.00 & 15.96 & $-$1.04 \\
BabyLM 2026 37M & GPT-2-705M teacher & 16.36 & 14.67 & $-$1.69 \\
BabyLM 2026 37M & Baby-Llama-58M (Llama+GPT-2) & 17.10 & 19.42 & $+$2.32 \\
BabyLM 2026 37M & Baby-Llama-58M (no KD) & 16.73 & 16.44 & $-$0.29 \\
\hline
\end{tabular}
\caption{Entity Tracking accuracy at the 128-token (tokenizer-declared) and 256-token (position-embedding maximum) context lengths. The 256-token setting is a diagnostic rather than the main protocol because all models were trained on 128-token slices. Most student rows move modestly, and the Britannica-vs-BabyLM ranking holds in every matched row at both context lengths. Some teacher rows, especially the 10M GPT-2 rows with learned absolute position embeddings, are more sensitive to the diagnostic setting.}
\label{tab:app-et256}
\end{table*}

\section{Training Token Counts}
\label{app:token-counts}

Table~\ref{tab:app-token-counts} reports the number of BPE tokens produced by each track-specific tokenizer on the corresponding training split. Word budgets count source words before inserting document boundary markers. Token counts include the \verb=<s>= and \verb=</s>= markers that are present during training.

\begin{table*}[tb]
\centering
\resizebox{\textwidth}{!}{
\begin{tabular}{llrrrr}
\hline
\textbf{Corpus} & \textbf{Budget} & \textbf{Source words} & \textbf{Boundary markers} & \textbf{BPE train tokens} & \textbf{Tokens / word} \\
\hline
Britannica & 10M & 10{,}000{,}066 & 18{,}362 & 14{,}216{,}855 & 1.42 \\
BabyLM 2026 & 10M & 10{,}000{,}092 & 2{,}331{,}378 & 19{,}518{,}278 & 1.95 \\
Britannica & 37M & 37{,}002{,}217 & 68{,}590 & 52{,}559{,}454 & 1.42 \\
BabyLM 2026 & 37M & 37{,}002{,}217 & 8{,}605{,}126 & 72{,}172{,}588 & 1.95 \\
\hline
\end{tabular}}
\caption{Training-token counts after corpus-specific BPE tokenization. The matched-word design follows the BabyLM convention, but it does not match the number of BPE tokens seen per epoch. In these splits, BabyLM 2026 yields about 1.37$\times$ as many training tokens as Britannica at both budgets, reflecting both corpus-specific BPE segmentation and BabyLM's many short source segments, which introduce far more document-boundary markers.}
\label{tab:app-token-counts}
\end{table*}

\section{Matched-Exposure Control}
\label{app:matched-exposure}

To test whether BabyLM~2026's greater number of token presentations explains the main 10M result, we retrain its seed-2023 Llama teacher, GPT-2 teacher, Llama+GPT-2 KD student, and no-KD student with the optimizer-step count capped at the corresponding Britannica run. All other settings, including the BabyLM split, tokenizer, architecture, effective batch size, warm-up, and loss, are unchanged. The caps are 3{,}472 steps for Llama, 2{,}604 for GPT-2, and 20{,}826 for each student. This exactly matches optimizer steps and schedule length and matches sequence presentations to within 0.03\%; because the BabyLM corpus contains more tokens, it yields 2.91 effective epochs for Llama and 4.37 for the other models rather than 4 and 6. This is a matched-exposure control, not a trim-matched corpus: the number of unique tokens available during training still differs. Table~\ref{tab:app-matched-exposure} reports the resulting scores.

\begin{table*}[tb]
\centering
\small
\begin{tabular}{lrrrrrrr}
\hline
\textbf{BabyLM 2026 model} & \textbf{Steps} & \textbf{Eff. ep.} & \textbf{BLiMP} & \textbf{Supp} & \textbf{EWoK} & \textbf{Entity} & \textbf{COMPS} \\
\hline
Llama-360M teacher & 3{,}472 & 2.91 & 68.91 & 57.59 & 50.60 & 20.52 & 51.51 \\
GPT-2-705M teacher & 2{,}604 & 4.37 & 69.14 & 53.66 & 50.12 & 14.24 & 51.65 \\
Baby-Llama-58M student (KD) & 20{,}826 & 4.37 & 70.22 & 56.39 & 50.95 & 16.40 & 51.66 \\
Baby-Llama-58M student (no KD) & 20{,}826 & 4.37 & 68.82 & 56.36 & 50.40 & 27.39 & 51.64 \\
\hline
\end{tabular}
\caption{Seed-2023 zero-shot results for the BabyLM~2026 10M matched-exposure control. Total optimizer steps and approximately total 128-token sequence presentations match the corresponding Britannica models.}
\label{tab:app-matched-exposure}
\end{table*}

Relative to the corresponding Britannica models in Table~\ref{tab:app10m}, the matched-exposure BLiMP differences (BabyLM minus Britannica) are $+7.46$, $+6.53$, $+5.53$, and $+5.81$ for the Llama teacher, GPT-2 teacher, KD student, and no-KD student, respectively; Supplement differences are $+8.78$, $+6.83$, $+9.79$, and $+8.91$. Entity Tracking differences are $-5.61$, $-11.14$, $-20.70$, and $+1.28$. Thus the acceptability advantage remains in all four families, while the Britannica Entity Tracking advantage remains in both teachers and the central KD comparison. The no-KD Entity Tracking row is the exception and reverses slightly, consistent with its reversal under seed 43 in Table~\ref{tab:corpusdelta}.

\section{BabyLM 2026 Split Details}
\label{app:babylm-splits}

Table~\ref{tab:app-babylm-splits} reports the manifests produced by the BabyLM 2026 split-preparation script. Word counts are counted before adding article/document boundary tokens. Because the preparation script keeps whole source documents, actual word counts can slightly exceed the requested targets.

\begin{table*}[tb]
\centering
\resizebox{\textwidth}{!}{
\begin{tabular}{lrrrrr}
\hline
\textbf{Split} & \textbf{Train words} & \textbf{Train docs} & \textbf{Val words} & \textbf{Val docs} & \textbf{Source rows consumed} \\
\hline
BabyLM 2026 10M & 10{,}000{,}092 & 1{,}165{,}689 & 2{,}027{,}481 & 236{,}073 & 1{,}401{,}762 \\
BabyLM 2026 37M & 37{,}002{,}217 & 4{,}302{,}563 & 1{,}003{,}428 & 116{,}690 & 4{,}419{,}253 \\
\hline
\end{tabular}}
\caption{BabyLM 2026 strict split manifests. Both splits are deterministic random document-level cuts from the same shuffled \texttt{Strict} source pool using seed 42.}
\label{tab:app-babylm-splits}
\end{table*}

\section{Paired Evaluation-Set Uncertainty}
\label{app:bootstrap-uncertainty}

For the default Llama+GPT-2 KD-student comparisons, we estimate evaluation-set uncertainty with a paired hierarchical bootstrap. Each of 10{,}000 replicates first resamples benchmark subtasks (the evaluation pipeline's UID groups) and then resamples paired items within each selected group. We recompute the official macro-average---the mean of subtask accuracies---and report the BabyLM~2026-minus-Britannica difference. Pairing ensures that both models are compared on the same resampled examples. These intervals are conditional on a trained checkpoint and are distinct from the across-seed variability reported in Table~\ref{tab:corpusdelta}. Table~\ref{tab:app-bootstrap} reports the full intervals.

\begin{table*}[tb]
\centering
\small
\begin{tabular}{llrrr}
\hline
\textbf{Budget} & \textbf{Training seed} & \textbf{$\Delta$BLiMP [95\% CI]} & \textbf{$\Delta$Supp [95\% CI]} & \textbf{$\Delta$Entity [95\% CI]} \\
\hline
10M & 2023 & $+7.07$ [$+3.01$, $+11.11$] & $+10.23$ [$+3.54$, $+16.49$] & $-22.27$ [$-24.96$, $-19.22$] \\
10M & 43   & $+8.19$ [$+4.05$, $+12.36$] & $+11.44$ [$+3.32$, $+19.58$] & $-17.00$ [$-19.77$, $-13.96$] \\
37M & 2023 & $+10.32$ [$+7.21$, $+13.75$] & $+5.74$ [$+0.47$, $+11.67$] & $-4.70$ [$-7.07$, $-2.36$] \\
37M & 43   & $+10.79$ [$+7.46$, $+14.53$] & $+8.42$ [$+2.29$, $+14.24$] & $-5.06$ [$-7.36$, $-2.89$] \\
\hline
\end{tabular}
\caption{Paired corpus effects and percentile 95\% hierarchical-bootstrap confidence intervals for the default KD students. Positive values favor BabyLM~2026; negative Entity Tracking values favor Britannica. All intervals exclude zero. Bootstrap seed: 2026.}
\label{tab:app-bootstrap}
\end{table*}
 
\end{document}